\documentclass[10pt,twocolumn,letterpaper]{article}

\usepackage[pagenumbers]{cvpr} 

\usepackage{graphicx}
\usepackage{amsmath}
\usepackage{amssymb}
\usepackage{booktabs}
\usepackage{bbm}

\usepackage[pagebackref,breaklinks,colorlinks]{hyperref}

\usepackage[capitalize]{cleveref}
\crefname{section}{Sec.}{Secs.}
\Crefname{section}{Section}{Sections}
\Crefname{table}{Table}{Tables}
\crefname{table}{Tab.}{Tabs.}

\def\eg{\emph{e.g.}} 

\def\ie{\emph{i.e.}}

\def\vs{\emph{v.s.}}
\def\wrt{w.r.t.} 

\def\etal{\emph{et al.}}

\def\cvprPaperID{510} 
\def\confName{CVPR}
\def\confYear{2022}

\begin{document}

\title{PlaneMVS: 3D Plane Reconstruction from Multi-View Stereo}

\author{Jiachen Liu$^{1,2}$\thanks{Work primarily done while interning at OPPO US Research Center} \quad Pan Ji$^{2}$\thanks{Corresponding author (peterji530@gmail.com)} \quad Nitin Bansal$^{2}$ \quad Changjiang Cai$^{2}$ \quad Qingan Yan$^{2}$ \quad Xiaolei Huang$^{1}$ \quad Yi Xu$^{2}$\\
$^1$The Pennsylvania State University \quad $^2$OPPO US Research Center, InnoPeak Technology, Inc.\\
}

\maketitle

\begin{abstract}
  We present a novel framework named PlaneMVS for 3D plane reconstruction from multiple input views with known camera poses. Most previous learning-based plane reconstruction methods reconstruct 3D planes from single images, which highly rely on single-view regression and suffer from depth scale ambiguity. In contrast, we reconstruct 3D planes with a multi-view-stereo (MVS) pipeline that takes advantage of multi-view geometry. We decouple plane reconstruction into a semantic plane detection branch and a plane MVS branch. The semantic plane detection branch is based on a single-view plane detection framework but with differences. The plane MVS branch adopts a set of slanted plane hypotheses to replace conventional depth hypotheses to perform plane sweeping strategy and finally learns pixel-level plane parameters and its planar depth map. We present how the two branches are learned in a balanced way, and propose a soft-pooling loss to associate the outputs of the two branches and make them benefit from each other. Extensive experiments on various indoor datasets show that PlaneMVS significantly outperforms state-of-the-art (SOTA) single-view plane reconstruction methods on both plane detection and 3D geometry metrics. Our method even outperforms a set of SOTA learning-based MVS methods thanks to the learned plane priors. To the best of our knowledge, this is the first work on 3D plane reconstruction within an end-to-end MVS framework. Source code: \url{https://github.com/oppo-us-research/PlaneMVS}.
\end{abstract}

\section{Introduction}
\label{sec:intro}

3D planar structure reconstruction from RGB images has been an important yet challenging problem in computer vision for decades. It aims to detect piece-wise planar regions and predict the corresponding 3D plane parameters from RGB images. The recovered 3D planes can be used in various applications such as robotics~\cite{taguchi2013point}, Augmented Reality (AR)~\cite{chekhlov2007ninja}, and indoor scene understanding~\cite{tsai2011real}.

\begin{figure}[!t]
\flushright
\vspace{-0.5cm}
\includegraphics[width=1.0\linewidth]{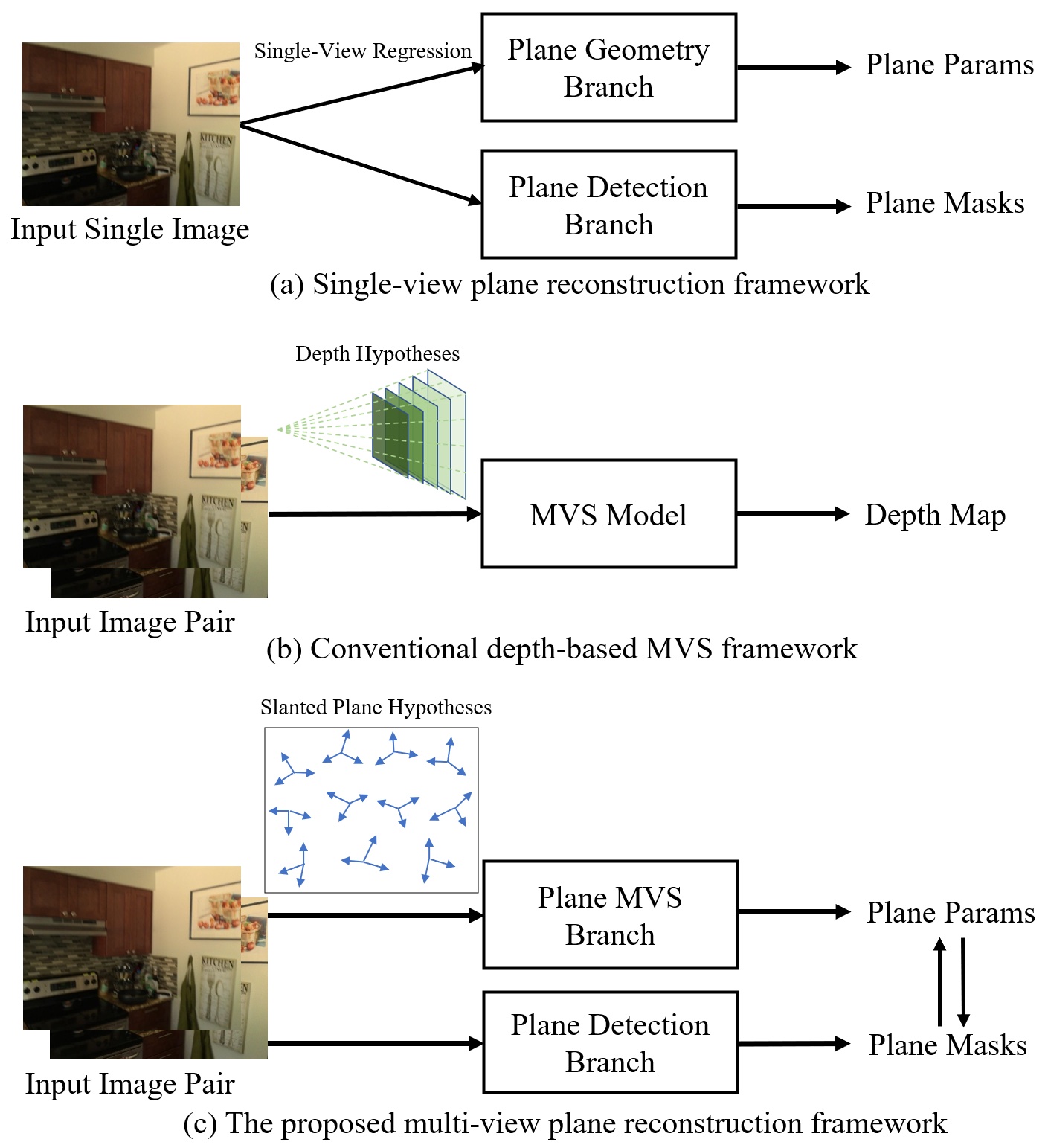}
\caption{Comparison among: (a) single-view plane reconstruction framework, (b) conventional depth-based MVS framework, (c) the proposed multi-view plane reconstruction framework. Our system employs slanted plane hypotheses for plane-sweeping to build the plane MVS branch, which interacts with the plane detection branch. The two branches can benefit from each other.}
\vspace{-0.6cm}
\label{fig:introduction}
\end{figure}

Traditional methods~\cite{furukawa2009manhattan,gallup2010piecewise,sinha2009piecewise} work well in certain cases but usually highly rely on some assumptions (\eg, Manhattan-world assumption~\cite{furukawa2009manhattan}) of the target scene and are thus not always robust in complicated real-world cases. Recently, some methods~\cite{liu2018planenet,yang2018recovering,liu2019planercnn,yu2019single,tan2021planetr} have been proposed to recover planes from single-view images based on Convolutional Neural Networks (CNNs). These methods could reconstruct 3D planes better in terms of completeness and robustness compared with traditional methods. However, all of them, albeit achieving reasonable results on 2D plane segmentation, attempt to recover 3D plane geometry from a single image, which is an ill-posed problem as it only relies on single-view regression for plane parameters and has ambiguity in depth scale recovery. Thus the recovered 3D planes from those methods are far from being accurate. The limitations of these methods motivate us to consider the possibility of reconstructing 3D planes from multiple views with CNNs in an end-to-end framework.

In contrast to reconstructing 3D geometry from single images, multi-view-stereo (MVS)~\cite{furukawa2009accurate} takes multiple images as input with known relative camera poses. MVS methods achieve superior performance on 3D reconstruction compared with single-view methods since the scale of a scene can be resolved by triangulating matched feature points on calibrated images~\cite{hartley2003multiple}. Recently, a few learning-based MVS methods~\cite{yao2018mvsnet,im2019dpsnet,gu2020cascade,yang2020cost} have been proposed and have achieved promising improvements for a wide range of scenes. While effective in reconstructing areas with rich textures, their pipelines would suffer from ambiguity in finding feature matches in the textureless area, which often belongs to planar regions. Besides, the generated depth map usually lacks smoothness as planar structures are not explicitly parsed. Some recent MVS approaches~\cite{kusupati2020normal,long2020occlusion,zhao2021confidence} propose to jointly learn the geometric relationship between depth and normal to capture local planarity. However, these methods usually estimate depth and normal separately, and only learn pixel-level planarity by enforcing the constraints by additional losses. Piece-wise planar structures, \eg, walls and floors, which usually indicate strong global geometric smoothness, are not well captured in these approaches.

In this work, as shown in Fig.~\ref{fig:introduction}, we take advantage of both sets of methods and propose to reconstruct planar structures in an MVS framework. Our framework consists of dual branches: a plane detection branch and a plane MVS branch. The plane detection branch predicts a set of 2D plane masks with their corresponding semantic labels of the target image. The plane MVS branch, which is our key contribution, takes posed target and source images as input. Inspired by the frontal plane sweeping formulation that is widely used in MVS pipelines, we propose a {\it slanted} plane sweeping strategy to learn the plane parameters without ambiguity. 
Specifically, instead of using a set of frontal plane hypotheses (\ie, depth hypotheses) for plane sweeping as in conventional MVS methods, we perform plane sweeping with a group of slanted plane hypotheses to build a plane cost volume for per-pixel plane parameter regression. 
\par
To associate the two branches, we present a soft-pooling strategy to get piece-wise plane parameters and propose a loss objective based on it to make the two branches benefit from each other. We apply learned uncertainties~\cite{kendall2017uncertainties} on different loss terms to train the multi-task learning system in a balanced way. Moreover, our system can generalize well in new environments with different data distributions. The results can be further improved with a simple but effective finetuning strategy without groundtruth plane annotations.

To the best of our knowledge, this is the first work that reconstructs planar structures in an end-to-end MVS framework. The reconstructed depth map takes advantage of multi-view geometry to resolve the scale ambiguity issue. It is much smoother geometrically compared with depth-based MVS schemes by parsing planar structures. Experimental results across different indoor datasets demonstrate that our proposed PlaneMVS not only significantly outperforms single-view plane reconstruction methods, but is also better than several SOTA learning-based MVS approaches.

\section{Related Work}
\label{sec:relatedwork}
\paragraph{Piece-wise planar reconstruction.}
Traditional plane reconstruction methods~\cite{furukawa2009manhattan,gallup2010piecewise,sinha2009piecewise} usually take a single or multiple images as input and detect the primitives such as vanishing points and lines as geometric cues to recover planar structures. Such methods make strong assumptions about the environment and often do not generalize well into various scenarios. Recent learning-based approaches~\cite{liu2018planenet,yang2018recovering,yu2019single,liu2019planercnn,tan2021planetr,qian2020learning} handle the plane reconstruction problem from a single image with Deep Neural Networks (DNNs) and achieve promising results. PlaneNet~\cite{liu2018planenet} proposes a multi-branch network to learn plane masks and parameters jointly. PlaneRecover~\cite{yang2018recovering} proposes to segment piece-wise planes with only groundtruth depth supervision but without any plane groundtruth. PlaneAE~\cite{yu2019single} and PlaneTR~\cite{liu2019planercnn} learn to cluster image pixels into piece-wise planes with bottom-up frameworks. Alternatively, PlaneRCNN~\cite{liu2019planercnn} takes advantage of a two-stage detection framework~\cite{he2017mask} to estimate plane segmentation and plane geometry in several parallel branches. Qian and Furukawa~\cite{qian2020learning} model the inter-plane relationships to further refine the initial planar reconstruction. However, although being possible to learn 2D plane segmentation with a single image, it is still challenging to learn accurate 3D plane geometry only with single-view regression. Most recently, Jin~\etal~\cite{jin2021planar} proposes a framework to jointly reconstruct planes and estimate camera poses from sparse views. In our work, we assume the camera poses are obtained from some SLAM systems, and design our plane detection branch based on PlaneRCNN~\cite{liu2019planercnn}, but learn plane geometry in a separate multi-view-stereo (MVS) branch.

\vspace{-0.4cm}

\paragraph{Multi-view stereo.}
Different from single-view depth estimation~\cite{eigen2015predicting,godard2019digging,tiwari2020pseudo,zou2020learning,ji2021monoindoor}, multi-view stereo transforms the depth estimation problem into triangulating corresponding points from a pair of posed images.~Thus, it could solve the scale ambiguity issue in the single-view case.~Traditional MVS approaches can be roughly categorized as voxel-based methods~\cite{kutulakos2000theory,seitz1999photorealistic}, point-cloud-based methods~\cite{furukawa2009accurate,lhuillier2005quasi} and depth-map-based methods~\cite{campbell2008using,galliani2015massively,tola2012efficient}.
\par
In recent years, some learning-based methods have been proposed and have shown superior robustness and generalizability. Volumetric methods such as~\cite{ji2017surfacenet,kar2017learning} aggregate multi-view information to learn a voxel representation of the scene.~However, they can only be applied to small-sized scenes due to high memory consumption of volumetric representation. For depth-based methods, MVSNet~\cite{yao2018mvsnet} utilizes an end-to-end framework to reconstruct the depth map of the reference image from multi-view input based on the plane-sweeping strategy. Some follow-up methods aim to achieve better accuracy-speed trade-off~\cite{yao2019recurrent,yu2020fast,wang2021patchmatchnet} or refine the depth map in a cascaded framework~\cite{gu2020cascade,yang2020cost}, or incorporate visibility as well as uncertainty into the framework~\cite{zhang2020visibility,luo2019p,xu2020pvsnet}. These depth map-based MVS approaches usually apply the fronto-parallel plane hypothesis for plane sweeping, aiming to learn pixel-level feature correspondences at correct depths. However, for textureless areas or repetitive patterns, it is challenging for the network to accurately match pixel-level features, thus making the inferred depth less accurate. Different from depth-map based MVS, Atlas~\cite{murez2020atlas} and NeuralRecon~\cite{sun2021neuralrecon} propose to learn a TSDF~\cite{curless1996volumetric} representation from posed images for 3D surface reconstruction which avoids multi-view depth fusion.
\par
Due to the matching ambiguity in textureless areas, some MVS works~\cite{birchfield1999multiway,gallup2007real,bleyer2011patchmatch} aim to model local planarity since textureless areas are usually planar. Traditionally, Birchfield and Tomasi~\cite{birchfield1999multiway} introduce slanted-plane with Markov Random Fields for stereo matching. Gallup~\etal~\cite{gallup2007real} first estimate dominant plane directions and warp along those planes based on plane sweeping. A few methods~\cite{bleyer2011patchmatch,xu2020planar,romanoni2019tapa} perform stereo patch matching in textureless regions based on iterative optimizations or probabilistic frameworks. For learning-based methods, derived from the idea of patchmatch stereo, a line of works~\cite{kusupati2020normal,long2020occlusion, zhao2021confidence} incorporate the geometric relationship between depth and surface normal into MVS framework. Although sharing high-level ideas, our work differs from these methods in several aspects. First, some work~\cite{long2020occlusion} segments piece-wise planes as an offline pre-processing step to generate smooth and consistent normals. However, we jointly learn plane segmentation and plane geometry within the proposed framework. Second, they usually learn depth and normal separately and apply loss objectives as extra constraints based on local planarity. In contrast, we directly learn to regress pixel-level plane parameters with a set of slanted plane hypotheses with the plane-sweeping strategy in one MVS pipeline, so the joint relationship between depth and normal is learned implicitly. Third, while those works aim to employ planar priors to assist multi-view stereo, our goal is to reconstruct piece-wise planar structures with an MVS framework.

\vspace{-0.1cm}

\section{Method}
\label{sec:method}

This section is organized as follows:~we first introduce our semantic plane detection branch in Sec.~\ref{sec:detection}, and present our plane MVS branch in Sec.~\ref{sec:planemvs}.~Then we describe the piece-wise plane reconstruction process in Sec.~\ref{sec:reconstruction}. Finally, we introduce our loss objectives in Sec.~\ref{sec:loss}.

\begin{figure*}[htp]
\centering
\vspace{-0.7cm}
\includegraphics[width=1.0\linewidth]{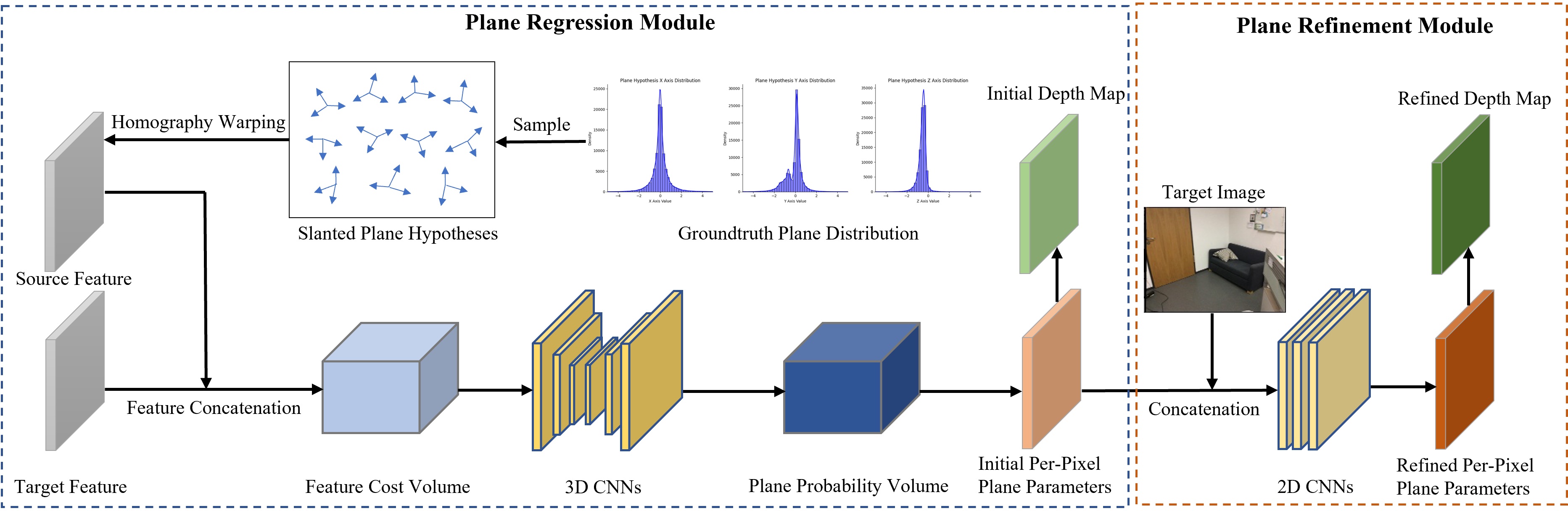}
\caption{The architecture of our proposed plane MVS head. It consists of a plane regression module to regress initial pixel-level plane parameters and a plane refinement module to refine the initial prediction. The key difference of our work from conventional MVS methods is that we apply slanted plane hypotheses for homography warping. The loss objectives during training are omitted for simplicity.}
\vspace{-0.4cm}
\label{fig:mvs_head}
\end{figure*}


\subsection{Plane detection}\label{sec:detection}

\par
{\bf An overview of PlaneRCNN.}
PlaneRCNN~\cite{liu2019planercnn} is one of the state-of-the-art single-view plane reconstruction approaches, which builds upon Mask-RCNN~\cite{he2017mask}. It designs several separate branches for estimating 2D plane masks and 3D plane geometry. It first applies FPN~\cite{lin2017feature} to extract a feature map, then adopts a two-stage detection framework to predict 2D plane masks $\mathcal{M}$. An encoder-decoder architecture processes the feature map to get a per-pixel depth map ${\bf D}$. Instance-level plane features from ROI-Align\cite{he2017mask} are passed into a plane normal branch to predict plane normals $\mathcal{N}$. They also design a refinement network to refine initial plane masks and a reprojection loss between neighboring views to enforce multi-view geometry consistency during training. With predicted $\mathcal{M}, \mathcal{N}$ and ${\bf D}$, the piece-wise planar depth map ${\bf D}_{p}$ can be reconstructed.

\par
{\bf Our semantic plane detection.}
Our detection head is based on PlaneRCNN~\cite{liu2019planercnn} but with several modifications. First, we remove all the geometry estimation modules including the plane normal prediction module and the monocular depth estimation module since 3D plane geometry is estimated by our MVS branch. Second, we also remove the plane refinement module and the multi-view reprojection loss used in PlaneRCNN to conserve memory. Additionally, since semantic information is helpful for scene understanding, as in Mask-RCNN~\cite{he2017mask}, we add semantic label prediction for each plane instance to get its semantic class. We will introduce the details on how we define and generate the semantic plane annotations in Sec.~\ref{sec:semantic_gt}. To summarize, for an input image with resolution ${H \times W}$, our plane detection head predicts a set of plane bounding boxes $\mathcal{B} = \{b_{1}, b_{2}, ..., b_{k}\}$, their confidence scores $\mathcal{S} = \{s_{1}, s_{2}, ..., s_{k}\}$ where $ s_{i} \in (0, 1) $, the binary plane masks $\mathcal{M} = \{{\bf m}_{1}, {\bf m}_{2}, ..., {\bf m}_{k}\}$ where $ {\bf m}_{i} \in \mathbb{R} ^ {H \times W}$, and their corresponding semantic labels $\mathcal{C} = \{c_{1}, c_{2}, ..., c_{k}\}$.

\subsection{Planar MVS}\label{sec:planemvs}
Next, we introduce our plane MVS head, which is our key contribution in this work.~Fig.~\ref{fig:mvs_head} shows the architecture of this branch, and we will present each part sequentially.

\par
{\bf Feature extraction.}
The 2D image feature extraction for the MVS head is shared with the plane detection head. Specifically, we obtain multi-scale 2D feature maps with $L=5$ levels from the FPN feature backbone. Here we only utilize the finest level feature $f_{0} \in \mathbb{R}^{\frac{1}{4}H \times \frac{1}{4}W \times C}$. To further balance the memory consumption and accuracy, we pass $f_{0}$ into a dimension-deduction layer and an average pooling layer to get reduced feature representation $f_{0}^{'} \in \mathbb{R}^{\frac{1}{8}H \times \frac{1}{8}W \times 
\frac{1}{2}C}$. $f_{0}^{'}$ serves as the feature map input of the MVS network. It is worth exploring whether using multiple levels of features would bring benefits, but that is not our current focus, and we leave it to future work.

\par
{\bf Differentiable planar homography.}
Previous MVS methods\cite{yao2018mvsnet,im2019dpsnet} propose to warp the source feature with fronto-parallel planes, \ie, depth hypotheses, to the target view. This is effective in associating the features from multiple views at the correct depth values of the target view. In our setup, the objective is to learn per-pixel plane parameters instead of depth. To this end, we propose to leverage slanted plane hypotheses for performing plane sweeping to learn per-pixel plane parameters with the MVS framework. The representation of differentiable homography using slanted plane hypotheses is the same as using depth hypotheses. The homography between two views at plane ${(\boldsymbol{n_{i}})}^{T}{\bf x} + e_{i} = 0$, where $\boldsymbol{n_{i}}$ is the plane normal and $e_{i}$ is the offset at pixel $i$ of the target view, can be represented as:
\begin{equation}\label{eqn:homography}
H_i(\boldsymbol{n_{i}}, e_{i}) \sim \boldsymbol{K} (\boldsymbol{R} - \frac{\boldsymbol{t}\boldsymbol{n_{i}}^{T}}{e_{i}})\boldsymbol{K^{-1}}\;,
\end{equation}
where symbol $\sim$ means ``equality up to a scale". $\boldsymbol{K}$ is the intrinsic matrix. $\boldsymbol{R}$ and $\boldsymbol{t}$ are the relative camera rotation and translation matrices between two views, respectively. Therefore it can be concluded that, without considering occlusion and object motion, the homography at pixel $i$ between two views is only determined by the plane $\boldsymbol{p_{i}}=\boldsymbol{n_{i}}^{T}/e_{i}$ with known camera poses. This perfectly aligns with our goal to learn 3D plane parameters with MVS. We can learn pixel-level plane parameters $\boldsymbol{p_{i}}=\boldsymbol{n_{i}}^{T} / e_{i}$, which is a non-ambiguous representation for a plane by employing slanted plane-sweeping in an MVS framework.

\par
{\bf Slanted plane hypothesis generation.}
One of the main differences of our framework from conventional MVS methods lies in the hypothesis representation. In depth-based MVS pipelines, their plane hypotheses are fronto-parallel~\wrt{}~the camera. Therefore, a set of one-dimensional depth hypotheses which cover the depth range of the target 3D space are sufficient for depth regression. However, in our work, we need a set of three-dimensional slanted plane hypotheses $\boldsymbol{n}^{T}/e$. Finding slanted plane hypotheses is a non-trivial task since the number of candidate planes that pass through a 3D point is infinite. We need to determine the appropriate hypothesis range for each dimension of $\boldsymbol{n}^{T}/e$. To this end, we randomly sample $10, 000$ training images and plot the distribution for every axis of groundtruth plane $\boldsymbol{n}^{T}/e$, which reflects the general distribution for plane parameters in various scenes. Then we select the upper and lower bounds for each axis by ensuring most groundtruth values lie within the selected range. We sample the hypotheses uniformly between the bounds along every axis. Please see details of the hypothesis range and number we choose in our supplementary material.

\par
{\bf Cost volume construction.}
After determining plane hypotheses, we warp the source feature map into the target view by Eq.~\eqref{eqn:homography}. For every slanted plane hypothesis, we concatenate the warped source feature and target feature to associate them, which can better keep the original single-view feature representation than applying distance metrics\cite{yao2018mvsnet}. Then we stack the features along the hypothesis dimension to build a feature cost volume $C$. Following \cite{yao2018mvsnet}, we utilize an encoder-decoder architecture with 3D CNN layers to regularize the feature cost volume. Finally, we use a single 3D CNN layer with softmax activation to transform the cost volume $C$ into a plane probability volume $U$.

\par
{\bf Per-pixel plane parameters inference and refinement.}
To make the whole system differentiable, following\cite{yao2018mvsnet}, soft-argmax is applied to get the initial pixel-level plane parameters. Given the plane hypothesis set $\mathcal{P}_{h} = $ \{$\boldsymbol{p_{0}}$, $\boldsymbol{p_{1}}$, ..., $\boldsymbol{p_{N-1}}$\}, 3D plane parameter $p_{i}$ at pixel $i$ can be inferred as:
\begin{equation}
\boldsymbol{p_{i}} = \sum_{j=0}^{N-1}\boldsymbol{p_{j}}\cdot U(\boldsymbol{p_{j}})\;,
\end{equation}
where $U(\boldsymbol{p_{i}})$ is the probability of hypothesis $\boldsymbol{p_{i}}$ at pixel $i$.

With soft-argmax, we can get an initial pixel-level plane parameter tensor ${\bf P} \in \mathbb{R} ^ {\frac{1}{8}H \times \frac{1}{8}W\times 3}$. We need to upsample it back to the original image resolution. We find that directly applying bilinear upsampling will lead to the over-smoothness issue. Here we adopt the upsampling method proposed by RAFT\cite{teed2020raft}. Specifically, for each pixel of ${\bf P}$, we learn a convex combination by first predicting an $8\times8\times3\times3$ grid, then applying weighted combination over the learned weights of its $3\times3$ coarse neighbors to get the upsampled plane parameters ${\bf P}^{'} \in \mathbb{R} ^ {H \times W\times 3}$. This upsampling approach better preserves the boundaries of planes and other details in the reconstructed planar depth map.

Following \cite{yao2018mvsnet}, we apply a refinement module, which aims to learn the residual of the initial plane parameters \wrt{} groundtruth. The upsampled initial plane parameters ${\bf P}^{'}$ is concatenated with the normalized original image $I_{t}$ as input to preserve image details, then passed into several 2D CNN layers to predict its residual $\bf{\delta P^{'}}$. Then we get the refined pixel-level plane parameters ${\bf P}_{r} = {\bf P}^{'} + \bf{\delta P^{'}}$, which is our final per-pixel plane parameters prediction.

\subsection{Planar depth map reconstruction}\label{sec:reconstruction}

In this subsection, we present how we associate the above two branches to make them benefit from each other. We also demonstrate how to get the piece-wise planar depth map as the final reconstructed plane representation.

\par
{\bf Plane instance-aware soft pooling.}
After getting per-pixel plane parameters and plane masks from the two branches, the natural question is, can we associate the two heads and make them benefit from each other? To this end, inspired by~\cite{yang2018recovering,yu2019single}, we design a soft-pooling operation and propose a loss supervision on the depth map. For a detected plane, we output its soft mask ${\bf m}_{s} \in \mathbb{R} ^{H \times W}$, where each element $\sigma_{i}$ at each pixel $i$ of ${\bf m}_{s}$ is the predicted foreground probability instead of a binary value $\in \{0, 1\}$ for differentiability. Then the instance plane parameter $\boldsymbol{p_{t}}$ can be computed by a soft-pooling operation with weighted averaging:
\vspace{-1mm}
\begin{equation}
\boldsymbol{p_{t}} = \frac{\sum_{i=1}^{N}\sigma_{i}\cdot \boldsymbol{p_{i}}}{\sum_{i=1}^{N}\sigma_{i}}\;.
\label{soft_pooling}
\end{equation}
\vspace{-1mm}
\par
Then the instance-level planar depth map can be reconstructed:
\vspace{-1mm}
\begin{equation}
{\bf D}_{i}=-\frac{\mathbbm{1}_{i}}{\boldsymbol{p_{t}}^T \boldsymbol{K^{-1}} {\bf x}_{i}}\;,
\label{plane2depth}
\end{equation}
where $\mathbbm{1}_{i}$ is an indicator variable to identify foreground pixels. A threshold of $0.5$ is applied on $\sigma_{i}$ to determine whether pixel $i$ is identified as foreground.~$\boldsymbol{K^{-1}}$ is the inverse intrinsic matrix and ${\bf x}_{i}$ is the homogeneous coordinate of pixel $i$.

\begin{table*}[htbp]
\centering
\vspace{-1mm}
\scalebox{0.65}{
\setlength{\tabcolsep}{6pt}{
\begin{tabular}{l|ccccccc|cccccc}
\hline
Method       &     \multicolumn{7}{c|}{Depth Metrics} & \multicolumn{5}{c}{Detection Metrics}\\
& AbsRel$\downarrow$ &SqRel$\downarrow$& RMSE$\downarrow$ & RMSE\_log$\downarrow$ & $\delta < 1.25$$\uparrow$ & $\delta < {1.25}^{2}$$\uparrow$ & $\delta < {1.25}^{3}$$\uparrow$ & AP$^{0.2m}$$\uparrow$ & AP$^{0.4m}$$\uparrow$ & AP$^{0.6m}$$ \uparrow$ & AP$^{0.9m}$$ \uparrow$  & AP$\uparrow$ & mAP$\uparrow$\\
\hline 
PlaneRCNN\cite{liu2019planercnn} & 0.164  & 0.068 & 0.284  & 0.186 & 0.780 & 0.953 & 0.989 & 0.310 & 0.475 & 0.526 & 0.546 & 0.554 & 0.452\\
MVSNet\cite{yao2018mvsnet}   & 0.105  & 0.040 & 0.232  &0.145 & 0.882 & 0.972 & 0.993 & - & - & - & - & - & -\\
DPSNet\cite{im2019dpsnet}   & 0.100  & 0.035 & 0.215  &0.135 & 0.896 & 0.977 & 0.994 & - & - & - & - & - & -\\
NAS\cite{kusupati2020normal}   & 0.098  & 0.035 & 0.213 & 0.134 & 0.905 & 0.979 & 0.994 & - & - & - & - & - & -\\
ESTDepth\cite{long2021multi} & 0.113 & 0.037 & 0.219 & 0.147 & 0.879 & 0.976 & 0.995 & - & - & - & - & - & - \\
PlaneMVS-pixel~(Ours) & 0.091  & 0.029 & 0.194  & 0.120 & 0.920 & 0.987 & 0.997 & 0.448 & 0.535 & 0.556 & 0.560 & \textbf{0.564} & \textbf{0.466}\\
PlaneMVS-final~(Ours) & \textbf{0.088}  & \textbf{0.026} & \textbf{0.186}  & \textbf{0.116} & \textbf{0.926} & \textbf{0.988} & \textbf{0.998} & \textbf{0.456} & \textbf{0.540} & \textbf{0.559} & \textbf{0.562} & \textbf{0.564} & \textbf{0.466}\\
\hline
\end{tabular}}}
\caption{Evaluation results for plane geometry and detection on ScanNet\cite{dai2017scannet} among different methods.}
\label{tab:compare_methods}
\vspace{-4mm}
\end{table*}

\par
{\bf Depth map representation and loss supervision.}
We can obtain a stitched depth map ${\bf D} \in \mathbb{R} ^{H \times W}$ for the image by filling the planar pixels with instance planar depth maps from Eq.~\eqref{plane2depth}. Since the learned pixel-wise plane parameters capture local planarity, we fill the non-planar pixels with the reconstructed pixel-wise planar depth map.
\par
Then we can design a soft-pooling loss $L_{sp}$ between the reconstructed depth map ${\bf D}$ and groundtruth depth map ${\bf D}^{*}$, with $L_1$ loss supervision as our soft-pooling loss $L_{sp}$:
\begin{equation}
L_{sp} = \left\|{\bf D} - {\bf D}^{*}\right\|_{1}\;.
\end{equation}
\par
By supervising the model with $L_{sp}$, because of Eq.~\eqref{soft_pooling}, the planar depth map is not only determined by the plane MVS head but also the plane detection head. In other words, the model is supposed to make the learned 2D plane segmentation and 3D plane parameters consistent with each other. During training, one module is able to get constraints from the other one's output. Note that although this loss shares similarity with \cite{yang2018recovering,yu2019single}, there are differences between their work and ours. PlaneRecover~\cite{yang2018recovering} applies a similar loss to assign pixels to different plane instances. PlaneAE~\cite{yu2019single} builds the loss on plane parameter instead of depth map and targets to improve instance-level parameters. In contrast, our soft-pooling loss is mainly designed for making possible interactions between 2D plane segmentation and 3D parameter predictions.

\subsection{Supervision with loss term uncertainty}\label{sec:loss}

Our supervisions have three parts:~the plane detection losses $L_{D}$, the plane MVS losses $L_{M}$, and the soft pooling loss $L_{sp}$.~$L_{D}$ includes two-stage classification and bounding box regression losses, and the mask loss in the $2^{\rm nd}$ stage. $L_{M}$ includes the loss built on initial per-pixel plane parameters and its reconstructed depth map, and the refined ones. For each term of $L_{M}$, we adopt masked $L_{1}$ loss which is only applied on the pixels with valid groundtruth. Since the goals of plane detection and plane MVS branch are distinct, we weight each loss term by its learned uncertainty as introduced in \cite{kendall2017uncertainties}. This is effective in our experiments and can outperform the results without applying uncertainty by a large margin. Our final loss objective can be written as:
\vspace{-1mm}
\begin{equation}
L = \sum_{i}^{N_{D}} \omega_{D_{i}}L_{D_{i}} + \sum_{j}^{N_{M}} \omega_{M_{j}}L_{M_{j}} + \omega_{sp}L_{sp}\;,
\end{equation}
\vspace{-1mm}
where $w$ is the learned uncertainty for each loss term.

\section{Experiments}
\label{sec:exps}

\subsection{Implementation details}\label{sec:implement}

We implement our framework in Pytorch~\cite{paszke2017automatic}. The SGD optimizer is applied with an initial learning rate of $3 \times 10^{-3}$ and a weight decay of $5 \times 10^{-4}$. The batch size is set to be $6$, and the model is trained end-to-end on $3$ NVIDIA 2080Ti GPUs for $10$ epochs on the ScanNet~\cite{dai2017scannet} benchmark. The learning rate decays to $3 \times 10^{-4}$ and $3 \times 10^{-5}$ at $7^{\rm th}$ and $9^{\rm th}$ epoch respectively. We re-implement plane detection module following ~\cite{liu2019planercnn} but with a publicly released implementation~\cite{massa2018mrcnn} of Mask-RCNN~\cite{he2017mask}. Following \cite{liu2019planercnn}, we initialize the weights with a detection model pretrained on COCO~\cite{lin2014microsoft}. The input image size is set to be $640 \times 480$ during training and testing. Since our batch size is relatively small, we freeze all the batch normalization~\cite{ioffe2015batch} layers of the plane detection head during training. We apply group normalization~\cite{wu2018group} as the normalization function in our plane MVS head.

\subsection{Training data generation}

\par
{\bf Semantic plane groundtruth generation.}\label{sec:semantic_gt}
To build our plane dataset with semantic labels, we first obtain and pre-process the raw rendered plane masks from~\cite{liu2019planercnn}, and get the 2D raw semantic maps from ScanNet~\cite{dai2017scannet}. Then we map the semantic labels from ScanNet to NYU40~\cite{silberman2012indoor}. We merge some semantically similar labels in NYU40 and finally pick $11$ labels that are likely to contain planar structures. We obtain the semantic label for each plane instance by projecting its mask onto the semantic map then performing a majority vote. If the voted result does not belong to any of the labels we select, we simply label the raw mask as non-planar and treat it as a negative sample during training and evaluation.

\par
{\bf View selection for MVS.}
We have to sample stereo pairs from ScanNet~\cite{dai2017scannet} monocular sequences and a stereo pair is considered appropriate if the images in the pair have a large enough camera baseline and sufficient overlap. In this work, we choose those stereo pairs as qualified ones if their relative translations lie between $0.05m$ and $0.15m$. We select $2$ views (a target and a source view) during training and testing. We believe adding more views could further improve performance, but that is outside the scope of this work.

\subsection{Datasets}

In our experiments, we use ScanNet~\cite{dai2017scannet} for training and evaluation. We further generalize our model to two other RGB-D indoor datasets, \ie, 7-Scenes~\cite{glocker2013real} and TUM-RGBD~\cite{sturm2012benchmark}, by testing with and without fine-tuning, to demonstrate the generalizability. Since the two datasets do not contain any plane groundtruths, we only evaluate the planar geometry metrics and show some qualitative results for plane detection on them. Due to space limit, here we only introduce how we use ScanNet. Please refer to our supplementary material for information on other datasets.

\begin{figure*}[htp]
\centering
\includegraphics[width=1.0\linewidth]{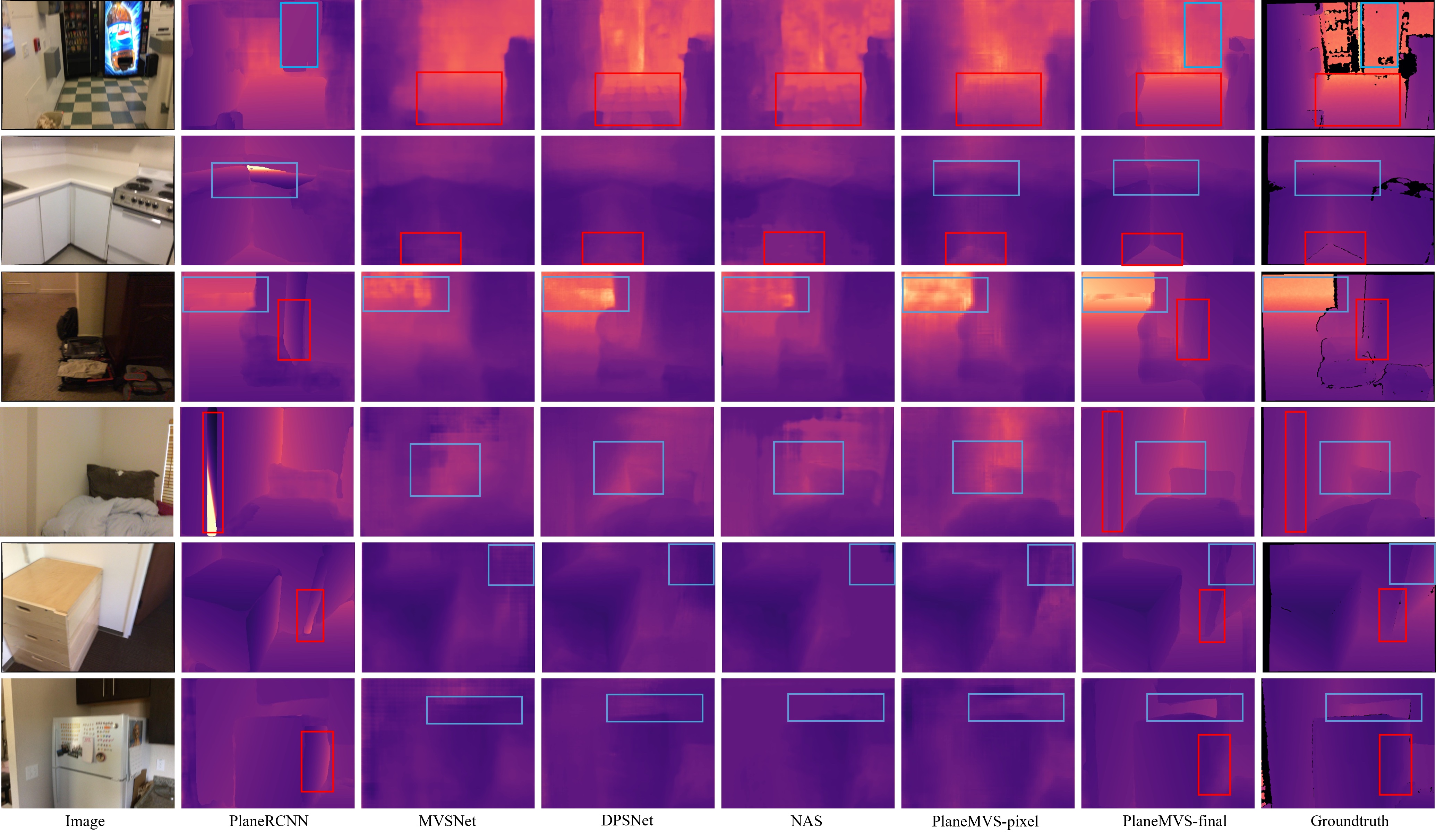}
\vspace{-0.5cm}
\caption{Qualitative results of reconstructed depth maps on ScanNet~\cite{dai2017scannet}. ``PlaneMVS-pixel" denotes the depth reconstructed from pixel-level plane parameters. ``PlaneMVS-final" denotes the final depth from the instance plane parameters after plane soft-pooling. Regions with salient differences between our results and others are highlighted with {\color{blue} blue} and {\color{red} red} boxes. Best viewed on screen with zoom-in.}
\vspace{-0.3cm}
\label{fig:qualitative}
\end{figure*}

\par
{\bf ScanNet.}
ScanNet~\cite{dai2017scannet} is a large indoor benchmark containing hundreds of scenes. We sample the training and testing stereo pairs from its official training and validation split, respectively. After pre-processing and filtering out the unqualified data following the steps in PlaneRCNN~\cite{liu2019planercnn}, we randomly subsample $20,000$ pairs for training. However, since the raw 3D meshes of ScanNet are not always complete, the rendered plane masks from meshes are noisy and inaccurate in quite a few images. This results in unconvincing plane detection evaluation if we directly test on those images. To this end, we manually pick 950 stereo pairs whose plane mask annotations are visually clean and complete from the original testing set for our evaluation.

\subsection{Evaluation metrics}

Following previous plane reconstruction methods~\cite{liu2018planenet,liu2019planercnn}, we mainly evaluate the plane reconstruction quality on average precision (AP) of plane detection with varying depth error thresholds $[0.2m, 0.4m, 0.6m, 0.9m]$, and the widely-used depth metrics~\cite{eigen2014depth}. Since we introduce plane semantics in our framework, we also evaluate the mean average precision (mAP)~\cite{lin2014microsoft} which couples semantic segmentation and detection as used in object detection papers.

\subsection{Comparison with state-of-the-arts}

\paragraph{Single-view plane reconstruction methods.}

\begin{figure}[!htp]
\flushleft
\includegraphics[width=1.0\linewidth]{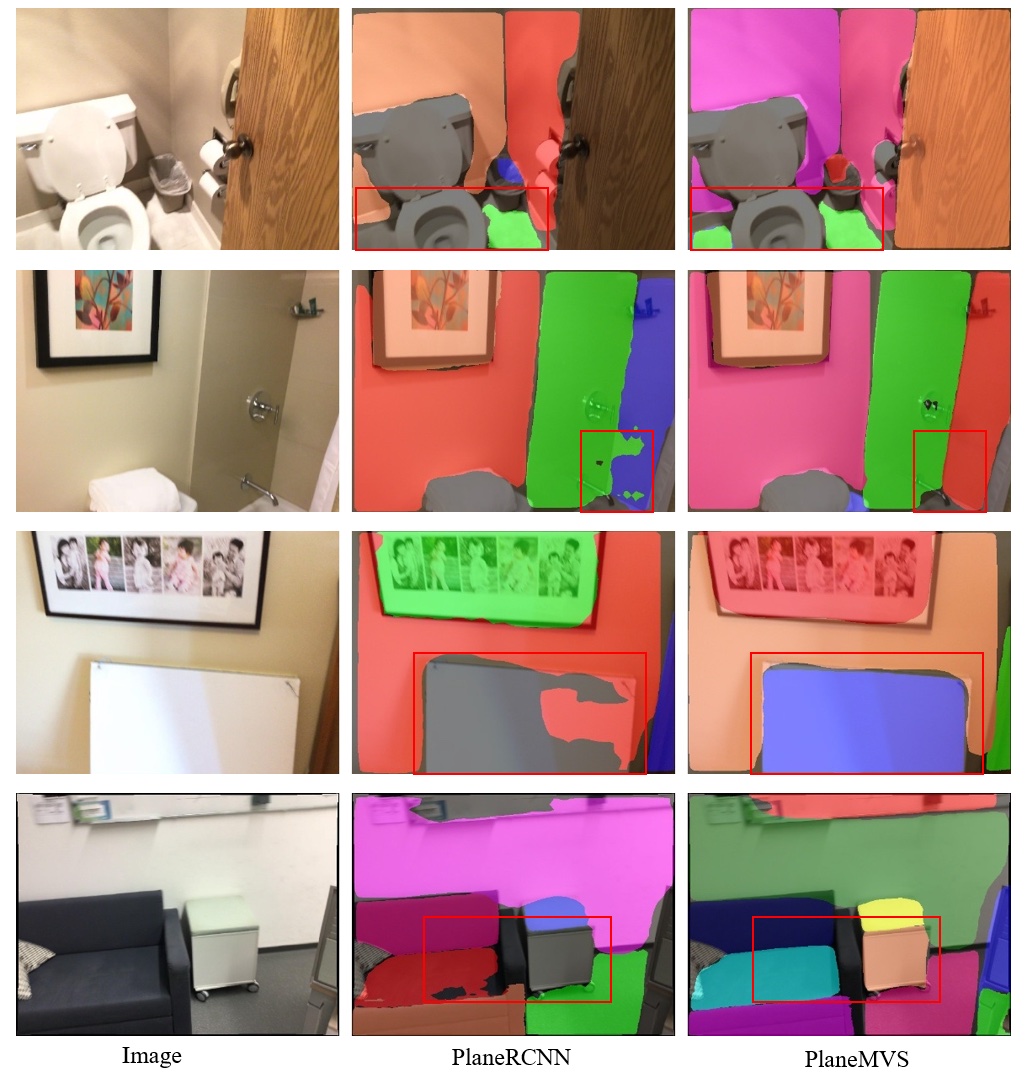}
\vspace{-0.5cm}
\caption{Qualitative results of plane detection on ScanNet~\cite{dai2017scannet}. The regions with salient differences between our method and PlaneRCNN are highlighted with {\color{red} red} boxes.}
\vspace{-0.1cm}
\label{fig:det_qualitative}
\end{figure}

We first compare our PlaneMVS with a SOTA single-view plane reconstruction method PlaneRCNN~\cite{liu2019planercnn}, which also serves as the baseline of our model. We test it on our re-implemented version with plane semantic predictions with the same training and testing data as ours. Tab.~\ref{tab:compare_methods} shows that our method outperforms PlaneRCNN in terms of both plane geometry and 3D plane detection by a large margin. As shown in Fig.~\ref{fig:qualitative}, PlaneRCNN does well in obtaining geometrically smooth planar depth maps, but their plane parameters are far from accurate (\eg, the $2^{nd}$ and $4^{th}$ row of Fig.~\ref{fig:qualitative}), which rely on single-view regression and suffer from the depth scale ambiguity issue (\eg, the $1^{st}$ and $3^{rd}$ row of Fig.~\ref{fig:qualitative}). For AP without considering depth, we also get considerable improvements benefiting from multi-task learning and the proposed soft-pooling loss. Although PlaneRCNN is a strong baseline, Fig.~\ref{fig:det_qualitative} clearly shows that our method better perceives plane boundaries, and our segmentation aligns better with 3D plane geometry. For mAP evaluation which considers plane semantic accuracy with detection, our method also outperforms PlaneRCNN by a nontrivial margin.
\par
{\bf Learning-based MVS methods.}
We also compare our method against several representative MVS methods. We select two representative depth-based MVS methods, MVSNet~\cite{yao2018mvsnet} and DPSNet~\cite{im2019dpsnet} since our MVS module shares similar network architecture with them. Besides, we also compare with NAS~\cite{kusupati2020normal} which aims to enforce depth-normal geometric consistency in MVS. We train and test these methods on our ScanNet data split with their released code for fair comparisons. We also compare with one of the state-of-the-art multi-view depth estimation methods ESTDepth~\cite{long2021multi}. From Tab.~\ref{tab:compare_methods}, our method clearly outperforms those MVS methods. Note that~\cite{long2021multi} is designed for temporally longer frames which may explain its possible performance drop when testing on two-views. The qualitative results in Fig.~\ref{fig:qualitative} clearly show that compared with conventional depth-based MVS methods, our ``PlaneMVS-pixel" results reconstructed from pixel-level plane parameters have shown more accurate depth, especially over textureless areas, which can be accredited to the proposed slanted plane hypothesis that learns planar geometry. By applying soft-pooling with detected plane masks (\ie, our ``PlaneMVS-final"), global geometric smoothness and sharper boundaries can be achieved over planar regions. The texture-copy issue in some cases (\eg, the $1^{st}$ row of Fig.~\ref{fig:qualitative}) of other methods can also be effectively avoided in ours. 

\subsection{Results on 7-Scenes}

\begin{table}[!htp]
\small
\centering
\scalebox{0.8}{
\setlength{\tabcolsep}{2pt}{
\begin{tabular}{l|ccc}
\hline
Method       &     AbsRel$\downarrow$ \qquad & $\delta < 1.25$$\uparrow$\\
\hline
PlaneRCNN~\cite{liu2019planercnn} & 0.221 & 0.640 \\
MVSNet~\cite{yao2018mvsnet} & 0.162 & 0.766 \\
DPSNet~\cite{im2019dpsnet} & 0.159 & 0.788 \\
NAS~\cite{kusupati2020normal} & 0.154 & 0.784 \\
ESTDepth~\cite{long2020occlusion} & \textbf{0.153} & 0.786 \\
Ours & 0.158  & \textbf{0.793} \\
\hline
Ours-FT & 0.113 & 0.890 \\
\hline
\end{tabular}}}
\caption{Reconstructed depth on 7-Scenes\cite{glocker2013real} dataset by different methods. ``Ours" means directly testing with ScanNet-pretrained model. ``Ours-FT" means testing with 7-Scenes-finetuned model.}
\label{tab:7-scenes}
\vspace{-4mm}
\end{table}

\begin{table*}[htp]
\centering
\vspace{-0.4cm}
\scalebox{0.7}{
\setlength{\tabcolsep}{6pt}{
\begin{tabular}{l|ccccccc|ccccc}
\hline
Method       &     \multicolumn{7}{c|}{Depth Metrics} & \multicolumn{5}{c}{Detection Metrics}\\
& AbsRel$\downarrow$ &SqRel$\downarrow$& RMSE$\downarrow$ & RMSE\_log$\downarrow$ & $\delta < 1.25$$\uparrow$ & $\delta < {1.25}^{2}$$\uparrow$ & $\delta < {1.25}^{3}$$\uparrow$ & AP$^{0.2m}$$\uparrow$ & AP$^{0.4m}$$\uparrow$ & AP$^{0.6m}$$ \uparrow$ & AP$^{0.9m}$$ \uparrow$  & AP$\uparrow$ \\
\hline 
Baseline & 0.170  & 0.074 & 0.305 & 0.200  & 0.746 & 0.944 & 0.990 & 0.288 & 0.458 & 0.519 & 0.545 & 0.551 \\
+ Soft-pooling loss   & 0.119  & 0.042 & 0.234 & 0.148 & 0.871 & 0.979 & 0.995 & 0.380 & 0.520 & 0.549 & 0.557 & 0.561 \\
+ Loss term uncertainty & 0.089  & 0.027 & 0.190 & 0.119 & 0.922 & 0.987 & 0.997 & 0.449 & 0.535 & 0.556 & 0.560 & 0.562 \\
+ Convex upsampling & \textbf{0.088}  & \textbf{0.026} & \textbf{0.186}  & \textbf{0.116} & \textbf{0.926} & \textbf{0.988} & \textbf{0.998} & \textbf{0.456} & \textbf{0.540} & \textbf{0.559} & \textbf{0.562} & \textbf{0.564}\\
\hline
\end{tabular}}}
\caption{Ablation study on the components of our proposed method.}
\vspace{-4mm}
\label{tab:ablation}
\end{table*}

We evaluate our approach on the 7-Scenes~\cite{glocker2013real} dataset to check its generalizability. Tab.~\ref{tab:7-scenes} shows that our method also significantly outperforms PlaneRCNN~\cite{liu2019planercnn}, and is better than or comparable with other MVS methods~\cite{yao2018mvsnet, im2019dpsnet, kusupati2020normal, long2021multi}. Since PlaneRCNN~\cite{liu2019planercnn} learns plane geometry from single views, the ability to generalize beyond the domain of training scenes is limited. However, our method benefits from multi-view geometry 
to learn multi-view feature correspondences and thus has superior generalizability on unseen data. We leave how we perform finetuning on 7-Scenes with only groundtruth depth to the supplementary material.

\subsection{Ablation study}
In this subsection, we evaluate the effectiveness of each proposed component (\ie, soft-pooling loss, loss term uncertainty, convex upsampling, and  slanted plane hypothesis). We leave some comparisons on hyper-parameters and settings to the supplementary material.

\begin{table}[!htp]
\small
\centering
\scalebox{0.8}{
{\setlength{\tabcolsep}{2pt}{
\begin{tabular}{l|ccccc}
\hline
Method       &     AbsRel$\downarrow$ & SqRel$\downarrow$ & $\delta < 1.25$$\uparrow$ & AP$^{0.2m}$ $\uparrow$ & AP$\uparrow$\\
\hline
Fronto-MVS & 0.094  & 0.033 & 0.917 & 0.433 & 0.548\\
Ours & \textbf{0.088}  & \textbf{0.026} & \textbf{0.926} & \textbf{0.456} & \textbf{0.564}\\
\hline
\end{tabular}}}}
{\footnotesize\caption{Ablation study: slanted v.s. fronto-parallel plane.}
\label{tab:depth-MVS}}
\vspace{-0.4cm}
\end{table}

\par
{\bf Soft-pooling loss.}
The soft-pooling loss is designed for coupling plane detection and plane geometry. From Tab.~\ref{tab:ablation}, it significantly improves both plane depth and 3D detection on all metrics. It helps the pixels within the same plane learns consistent plane parameters which benefits the plane geometry. For plane detection, as shown in Fig.~\ref{fig:det_qualitative}, our detected planes also align better with 3D plane geometry, especially over plane boundaries.

\par
{\bf Training with loss term uncertainty.}
Tab.~\ref{tab:ablation} shows that by weighting each loss term with learned uncertainty, our model can get further improvement on plane geometry and 3D detection. There are two possible reasons. First, our model has different branches with multiple losses for 2D or 3D objectives. Setting each loss with the same weight may not make them converge smoothly. Second, as introduced in Sec.~\ref{sec:implement}, the plane detection head is initialized with a COCO-pretrained model. But for the MVS head, we train it from scratch. Thus the learning procedures for the two branches may be imbalanced if we do not adaptively change the weights for each term. After adopting the learnable uncertainty, the weights of different terms are automatically tuned during training, which has brought great benefits.

\par
{\bf Convex upsampling.}
We analyze the effect of applying the convex combination upsampling to replace bilinear upsampling. As shown in Tab.~\ref{tab:ablation}, we get considerable improvement. The learned convex upsampling better keeps fine-grained details than bilinear upsampling.

\par
{\bf Slanted plane hypothesis.}
We conduct an additional ablation study, \ie, replacing our slanted plane hypothesis with the frontal-parallel plane hypothesis, using the same network architecture. We also apply convex upsampling and loss-term uncertainty for fair comparison. We employ the least-squares algorithm to fit planes with the predicted per-pixel depth map and plane masks, and then transform the plane parameters to planar depth maps. As shown in Tab.~\ref{tab:depth-MVS}, our proposed method outperforms the `Fronto-MVS' baseline on both 3D plane detection and depth metrics. Besides, our model learns plane parameters in an end-to-end manner instead of fitting planes as a post-processing step. This verifies the effectiveness of the proposed slanted plane hypothesis~\wrt{} fronto-parallel hypothesis.

\section{Conclusion and Future Work}
\label{sec:cons}

In this work, we propose PlaneMVS, a deep MVS framework for multi-view plane reconstruction. Based on our proposed slanted plane hypothesis for plane-sweeping, 3D plane parameters can be learned by deep MVS in an end-to-end manner. We also couple the plane detection branch and the plane MVS branch with the proposed soft pooling loss. Compared with single-view methods, our system can reconstruct 3D planes with significantly better accuracy, robustness, and generalizability. Without sophisticated designs, our system even outperforms several state-of-the-art MVS approaches. Please refer to our supplementary material for more results, discussions and potential limitations.
\par
There are a few directions worth exploring in the future. First, recent advanced designs of deep MVS systems~\cite{gu2020cascade,yang2020cost,zhang2020visibility,wang2021patchmatchnet} could be incorporated to further improve MVS reconstruction. Second, temporal information from videos (beyond two frames as we are currently using) can be exploited to achieve temporally coherent plane reconstruction, such that consistent single-view predictions could be fused into a global 3D model of the entire scene. 
\vspace{-0.4cm}
\paragraph{Acknowledgement.} The research of J. Liu and X. Huang was supported in part by the NSF award \#1815491.

\newpage

\section{Supplementary Material}

\subsection{Hypothesis selection for slanted planes}

\begin{figure*}[htp]
\centering
\includegraphics[width=0.7\linewidth]{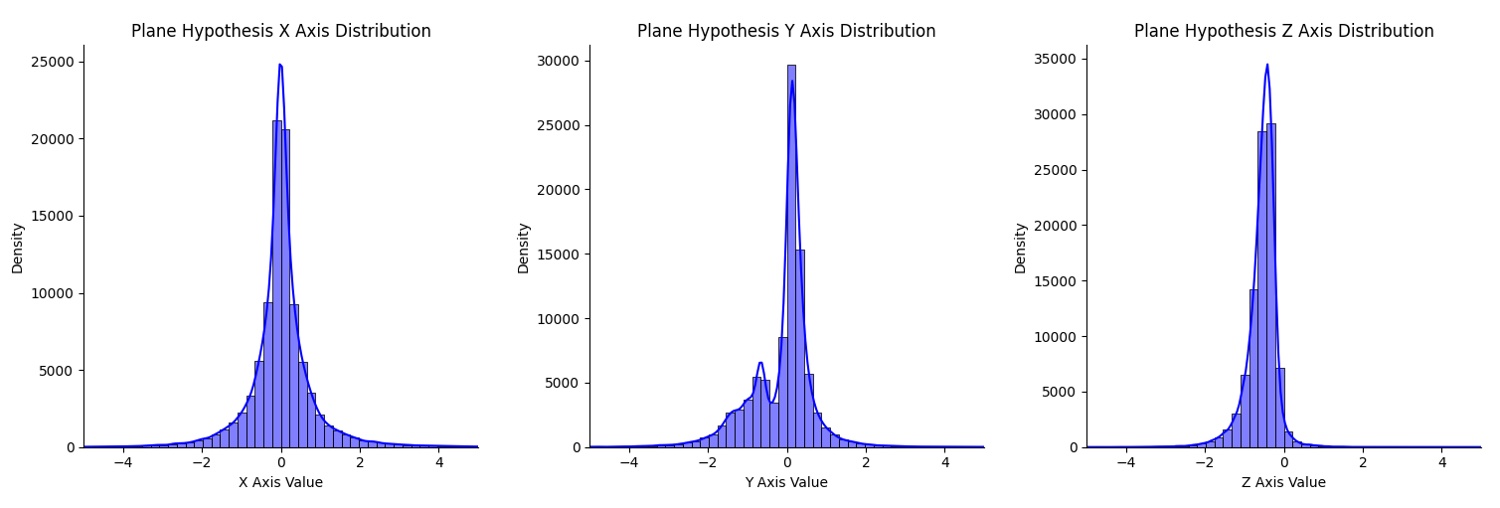}
\caption{Plane hypothesis distribution of the three axes.}
\label{fig:plane_dist}
\end{figure*}

Fig.~\ref{fig:plane_dist} shows the distribution of the three axes of plane $\boldsymbol{n}^{T}/e$ sampled from $10, 000$ training images. Based on the distribution, we select $(-2, 2)$, $(-2, 2)$, $(-2, 0.5)$ as the range of $x, y, z$ axis for $\boldsymbol{n}^{T}/e$, respectively, to ensure at least $95\%$ of the groundtruth planes lie within the ranges. Since our plane hypothesis is a three-dimensional vector, the computational cost of the cost volume is cubic \wrt{}~the number of hypothesis per axis. To reach a balance between accuracy and memory consumption, we sample 8 hypotheses uniformly along every axis and finally have $N=8^{3}=512$ plane hypotheses in total.

\subsection{Semantic classes on ScanNet}

\begin{figure}[!htp]
\flushleft
\includegraphics[width=1.0\linewidth]{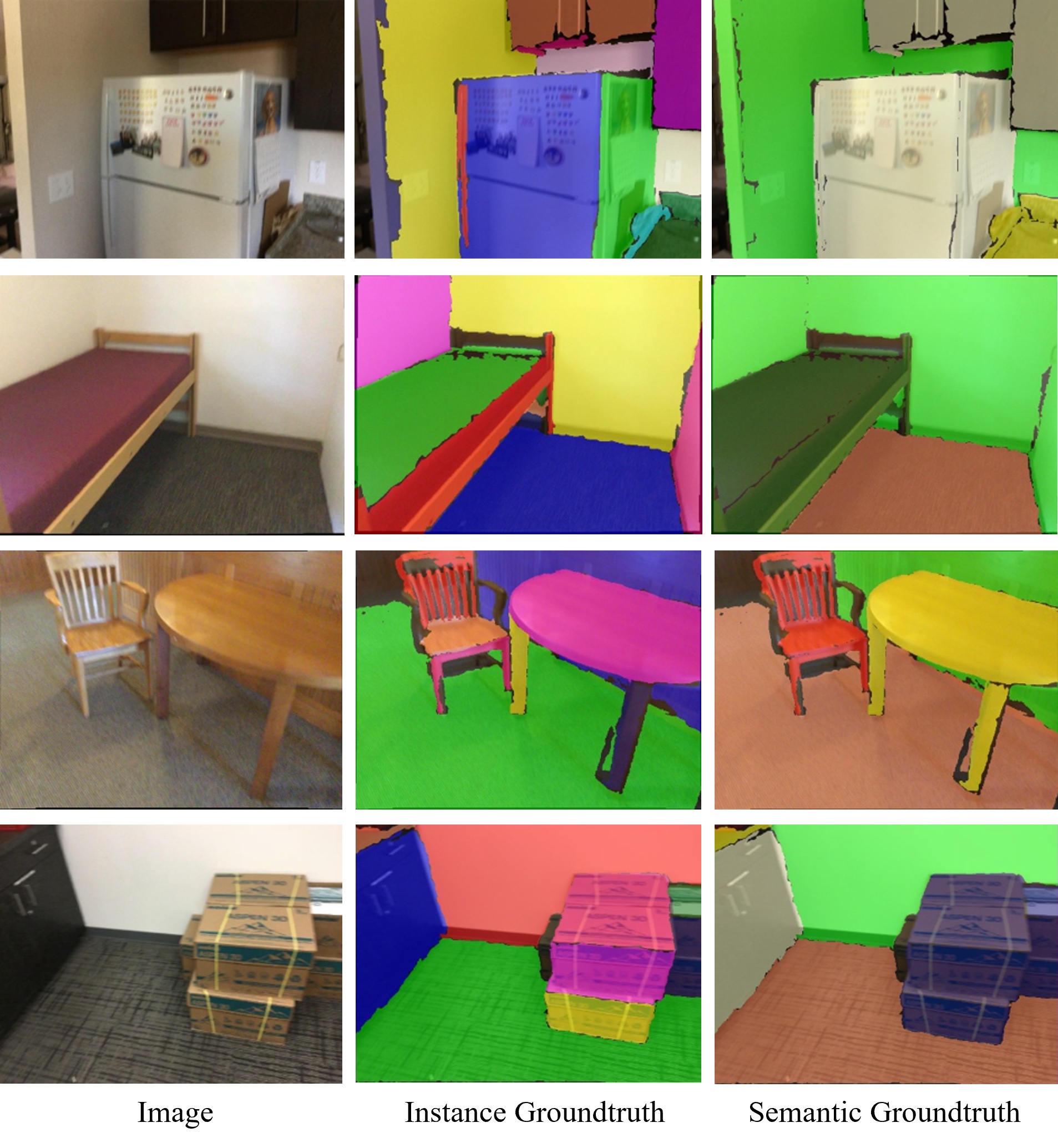}
\vspace{-0.3cm}
\caption{Examples of planar semantic and instance groudtruths on ScanNet~\cite{dai2017scannet}. Different colors represent different plane instances ($2^{nd}$ column) or semantic categories ($3^{rd}$ column).}
\label{fig:plane_gt_qualitative}
\end{figure}

After merging the semantically-similar categories in NYU40~\cite{silberman2012indoor} labels, we pick 11 classes: wall, floor, door, chair, window, picture, desk \& table, bed \& sofa, monitor \& screen, cabinet \& counter, box \& bin, which are likely to contain planar structures in indoor scenes. Please refer to Fig.~\ref{fig:plane_gt_qualitative} for some visualization examples of the generated planar instance and semantic groundtruth from ScanNet\cite{dai2017scannet}.

\subsection{Benchmark setup}

\par
{\bf 7-Scenes.} 7-Scenes~\cite{glocker2013real} collects posed RGB-D camera frames of seven indoor scenes. We sample stereo pairs in the same manner as in ScanNet~\cite{dai2017scannet} and follow the official split to get finetuning and evaluation data. We finally have $26,358$ pairs for finetuning and $15,508$ pairs for evaluation.
\par
{\bf TUM-RGBD.} TUM-RGBD~\cite{sturm2012benchmark} is an indoor RGB-D monocular SLAM dataset with calibrated cameras. We randomly select 4 scenes (\ie, fr1-desk, fr1-room, fr1-desk2, fr3-long-office-household) with $5,013$ pairs for finetuning and 2 scenes (\ie, fr2-desk, fr3-long-office-household-validation) containing $4,817$ pairs for evaluation.

\subsection{Results on 7-Scenes and TUM-RGBD}

\begin{figure*}[htp]
\centering
\vspace{-1mm}
\includegraphics[width=1.0\linewidth]{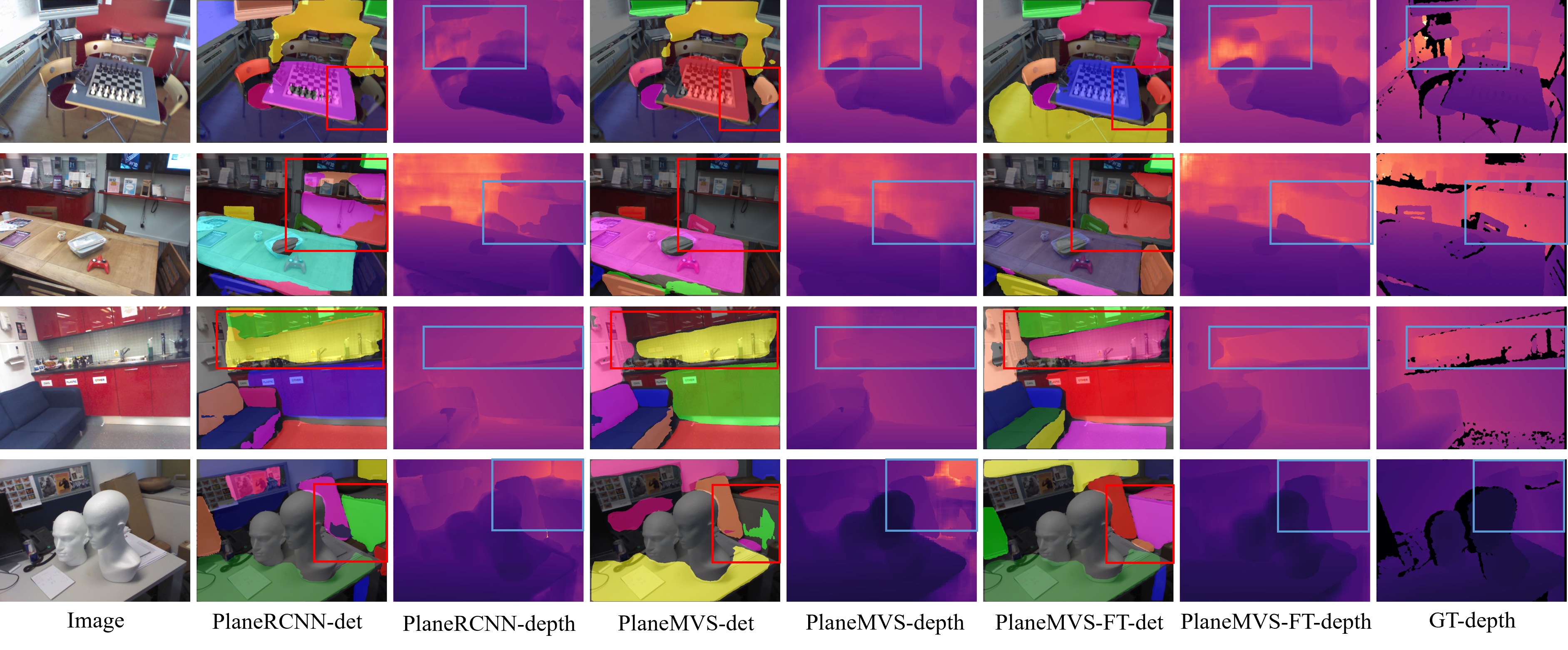}
\caption{The plane reconstruction results on 7-scenes~\cite{glocker2013real} among different methods. ``FT'' denotes ``finetuned'' and ``det'' is short for ``detection''. Regions with salient differences are highlighted with {\color{blue} blue} and {\color{red} red} boxes.}
\label{fig:7scenes_qualitative}
\end{figure*}

We have discussed how we deal with 7-Scenes and have demonstrated its quantitative results in the main paper. Here we introduce our simple but effective strategy to perform finetuning with only groundtruth depth. We first generate pseudo groundtruths of plane masks by getting the predictions with the ScanNet-pretrained model on the testing images. Then we train our model without plane parameter losses but maintain other losses. We simply set each loss weight to $1$ instead of adopting the loss term uncertainty during finetuning since we find it cannot bring much improvement. We finetune the model for $5$ epochs. The planar depth gets much improved and we find that the plane detection results also tend to be visually better, which may be accredited to multi-task training and our soft-pooling loss to associate 2D with 3D. The same applies to the TUM-RGBD~\cite{sturm2012benchmark} dataset. Some qualitative examples of 7-Scenes are shown in Fig.~\ref{fig:7scenes_qualitative}.

\par
\begin{figure*}[htp]
\centering
\vspace{-1mm}
\includegraphics[width=1.0\linewidth]{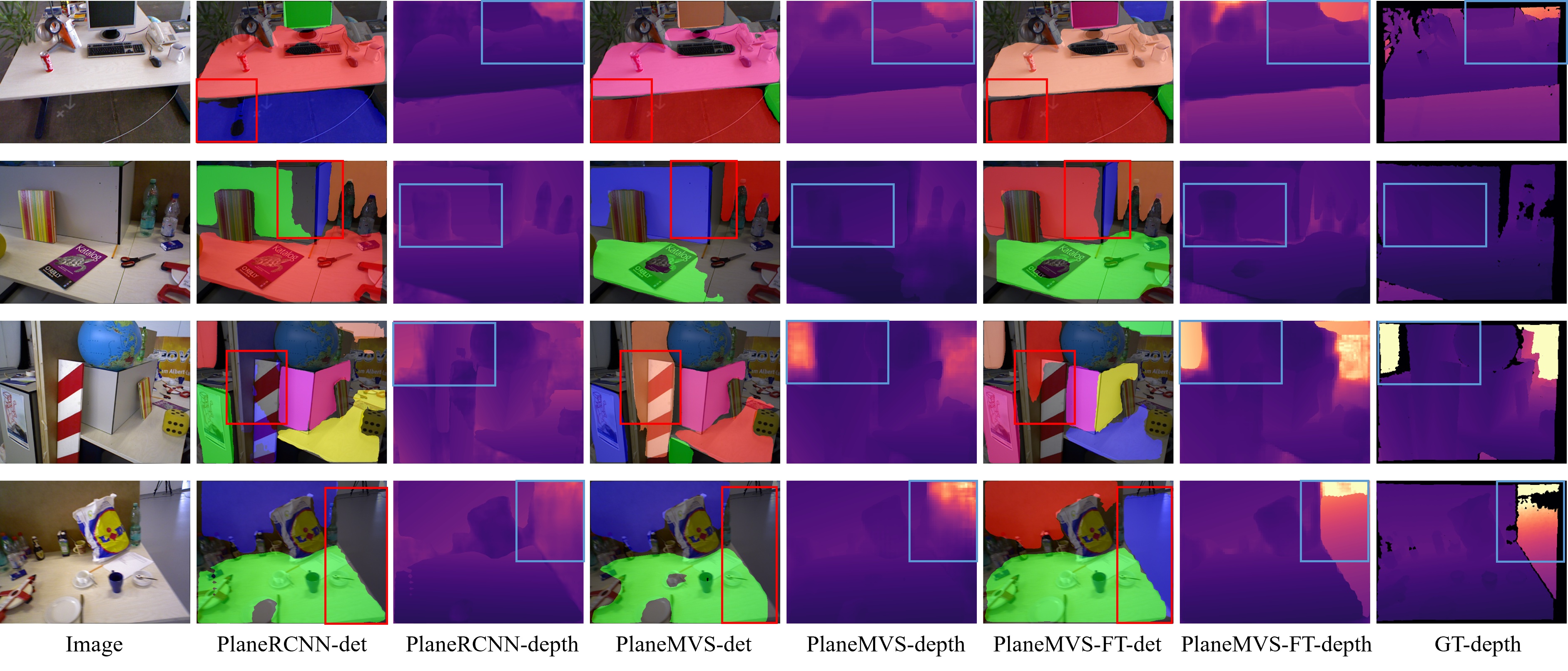}
\caption{The plane reconstruction results on TUM-RGBD~\cite{sturm2012benchmark} among different methods. Regions with salient differences are highlighted with {\color{blue} blue} and {\color{red} red} boxes.}
\label{fig:tum_qualitative}
\end{figure*}

As shown in Tab.~\ref{tab:tum-rgbd} and Fig.~\ref{fig:tum_qualitative}, similar to 7-Scenes, our approach generalizes much better on TUM-RGBD compared with PlaneRCNN~\cite{liu2019planercnn}, thanks to the learned multi-view geometric relationship. By performing the proposed finetuning strategy, the results get further improved on both 3D planar geometry and 2D planar detection.

\begin{table}[htp]
\centering
\vspace{-1mm}
\scalebox{0.8}{
\setlength{\tabcolsep}{2pt}{
\begin{tabular}{l|ccc}
\hline
Method       &     AbsRel$\downarrow$ & SqRel$\downarrow$ & $\delta < 1.25$$\uparrow$\\
\hline
PlaneRCNN\cite{liu2019planercnn} & 0.243 & 0.105 & 0.655 \\
Ours & 0.143  & 0.07 & 0.795 \\
Ours-FT & \textbf{0.120}  & \textbf{0.054} & \textbf{0.851} \\
\hline
\end{tabular}}}
\caption{Reconstructed depth on TUM-RGBD dataset~\cite{sturm2012benchmark} of different methods. ``Ours" means directly testing with the ScanNet-pretrained model. ``Ours-FT" means testing with the TUM-RGBD-finetuned model.}
\label{tab:tum-rgbd}
\vspace{-1mm}
\end{table}

\subsection{More Ablation studies}

In this section, we discuss the impact of applying different hyper-parameters or settings in our experiments. Then we show qualitative examples on the two components of our proposed method to intuitively demonstrate their effects.

\subsubsection{Hyper-parameters and settings}

\par
{\bf Plane hypothesis range.}
We first study the effect of the plane hypothesis range we set. We compare the results of different hypothesis ranges while keeping the hypothesis number $N$ unchanged: (i) use the same range of $(-2, 2)$ for the $x, y, z$ axes; (ii) broaden the range to $(-2.5, 2.5)$; (iii) shorten the range to  $(-1.75, 1.75)$; (iv) employ the same range of $(-2, 2)$ for the $x, y$ axes and a different range of $(-2, 0.5)$ for the $z$ axis. As shown in Tab.~\ref{tab:hypo_range}, setting (iv), which serves as our default setting, achieves the best result. The performance drops when using the same range for all axes as (i), since $z$ values mainly distribute between $(-2, 0.5)$. Using a broader range, \eg~(i) and (ii), covers some marginal values but decreases the density of the plane hypothesis, thus leading to less accurate results. In setting (iv), although shortening ranges can increase the hypothesis density, some non-negligible groundtruth values are not well covered, thus also leading to worse results.

\begin{table}[htp]
\centering
\vspace{-1mm}
\scalebox{0.8}{
\setlength{\tabcolsep}{2pt}{
\begin{tabular}{l|cc}
\hline
Hypos range       & AbsRel$\downarrow$ & $\delta < 1.25$$\uparrow$ \\
\hline
(-2, 2) for x,y,z & 0.093 & 0.920 \\
(-1.75, 1.75) for x,y,z & 0.094 & 0.921 \\
(-2.5, 2.5) for x,y,z & 0.096 & 0.919 \\
(-2, 2) for x,y; (-2, 0.5) for z & \textbf{0.088}  & \textbf{0.926}\\
\hline
\end{tabular}}}
\caption{Ablation study on the range of slanted plane hypothesis.}
\label{tab:hypo_range}
\vspace{-1mm}
\end{table}

\par
{\bf Plane hypothesis number.}
When keeping the plane hypothesis range constant, varying hypothesis number $N$ changes the hypothesis density. We test our model using $6, 8, 10$ hypotheses per axis, \ie, $N = 216, 512$ and $1,000$ respectively. The results are listed in Tab.~\ref{tab:hypo_num}. As expected, in general, the higher density we set, the better geometry performance we achieve. The performance gaps among different numbers are small, which demonstrates that our model is robust to these hyper-parameters to some extent. Note that using $N=1,000$ will substantially increase the memory consumption. So we choose $N=512$ in our default setting.

\begin{table}[htp]
\centering
\vspace{-1mm}
\scalebox{0.8}{
\setlength{\tabcolsep}{2pt}{
\begin{tabular}{l|cc}
\hline
Hypos number per axis  & AbsRel$\downarrow$ & $\delta < 1.25$$\uparrow$\\
\hline 
6 hypos (216 in total) & 0.091  & 0.924\\
8 hypos (512 in total) & \textbf{0.088}  & 0.926\\
10 hypos (1,000 in total) & \textbf{0.088} & \textbf{0.927}\\
\hline
\end{tabular}}}
\caption{Ablation study on plane hypothesis number.}
\label{tab:hypo_num}
\vspace{-1mm}
\end{table}

\begin{table}[htp]
\centering
\vspace{-1mm}
\scalebox{0.8}{
\setlength{\tabcolsep}{2pt}{
\begin{tabular}{l|cc}
\hline
Method       &     AbsRel$\downarrow$ & $\delta < 1.25$$\uparrow$\\
\hline
Pixel-planar w/o pooling & 0.091 & 0.920 \\
Pooling with predicted masks & 0.088  & 0.925 \\
Soft-pooling with predicted masks & 0.088  & 0.926 \\
Pooling with groundtruth masks & \textbf{0.087} & \textbf{0.932} \\
\hline
\end{tabular}}}
\caption{Ablation study on plane instance pooling with plane masks during testing.}
\label{tab:soft_pool}
\vspace{-1mm}
\end{table}

\par
{\bf Plane instance-aware soft pooling.}
We now evaluate the recovered depths among different pooling strategies reflecting the efficacy of plane detection on the learned 3D planar geometry. As shown in Tab.~\ref{tab:soft_pool}, when evaluating the depth reconstructed from pixel-level plane parameters, it underperforms the results with plane instance pooling since the generated depth maps cannot capture piece-wise planarity. The result improves when we apply hard-pooling with predicted plane masks over the pixel-level plane parameters. Applying soft-pooling weighted with pixel-level probability further brings a minor improvement since the probability reflects the confidence of a pixel belonging to a plane instance. Finally, we use groundtruth plane masks to perform pooling, which represents the upper bound of the impact of plane detection on geometry. As expected, it achieves the best result among the settings. Since groundtruth plane masks are not available during testing, we always apply the soft-pooling with predicted masks in other experiments.

\par
{\bf Depth on planar region.}
We further compare the reconstructed depth over only planar regions \vs{}~the whole image. Specifically, we conduct experiments only evaluating depth on the pixels that belong to any of the groundtruth planes. As shown in Tab.~\ref{tab:planar_depth}, compared with the depth over the whole image, the quantitative result over planar regions is better, no matter whether plane-instance-pooling is applied or not. This demonstrates that our proposed method's geometry improvement mainly comes from the pixels of planar regions, which conforms to our initial motivation and objective.

\begin{table}[htp]
\centering
\vspace{-1mm}
\scalebox{0.8}{
\setlength{\tabcolsep}{2pt}{
\begin{tabular}{l|cc}
\hline
Method & AbsRel$\downarrow$ & $\delta < 1.25$$\uparrow$\\
\hline
Depth over whole image w/o pooling & 0.091 & 0.920\\
Depth over planar region w/o pooling & 0.086  & 0.929\\
Depth over whole image & 0.088  & 0.926\\
Depth over planar region & \textbf{0.081}  & \textbf{0.938}\\
\hline
\end{tabular}}}
\caption{Ablation study on the evaluations over planar region.}
\label{tab:planar_depth}
\vspace{-1mm}
\end{table}

\par
{\bf Training dataset scale.}
In our default setting, we only sample $20,000$ stereo pairs for training. To analyze the impact of the scale of training data, we sample a larger training set with $66,000$ stereo pairs from the same scene split but keep the evaluation split unchanged. As shown in Tab.~\ref{tab:data_scale}, our performance can be further improved with more training data on both plane detection and geometry metrics.

\begin{table}[htp]
\centering
\vspace{-1mm}
\scalebox{0.8}{
\setlength{\tabcolsep}{2pt}{
\begin{tabular}{l|cccc}
\hline
Dataset Scale       &   AbsRel$\downarrow$ & $\delta < 1.25$$\uparrow$ & AP$^{0.2m}$ $\uparrow$ & AP$\uparrow$\\
\hline
20,000 training pairs & 0.088  & 0.926 & 0.456 & 0.564\\
66,000 training pairs & \textbf{0.082}  & \textbf{0.934} & \textbf{0.470} & \textbf{0.570}\\
\hline
\end{tabular}}}
\caption{Ablation study on the scale of training dataset.}
\label{tab:data_scale}
\vspace{-1mm}
\end{table}

\subsubsection{Qualitative ablation analysis}

\begin{figure}[htp]
\flushright
\includegraphics[width=1.0\linewidth]{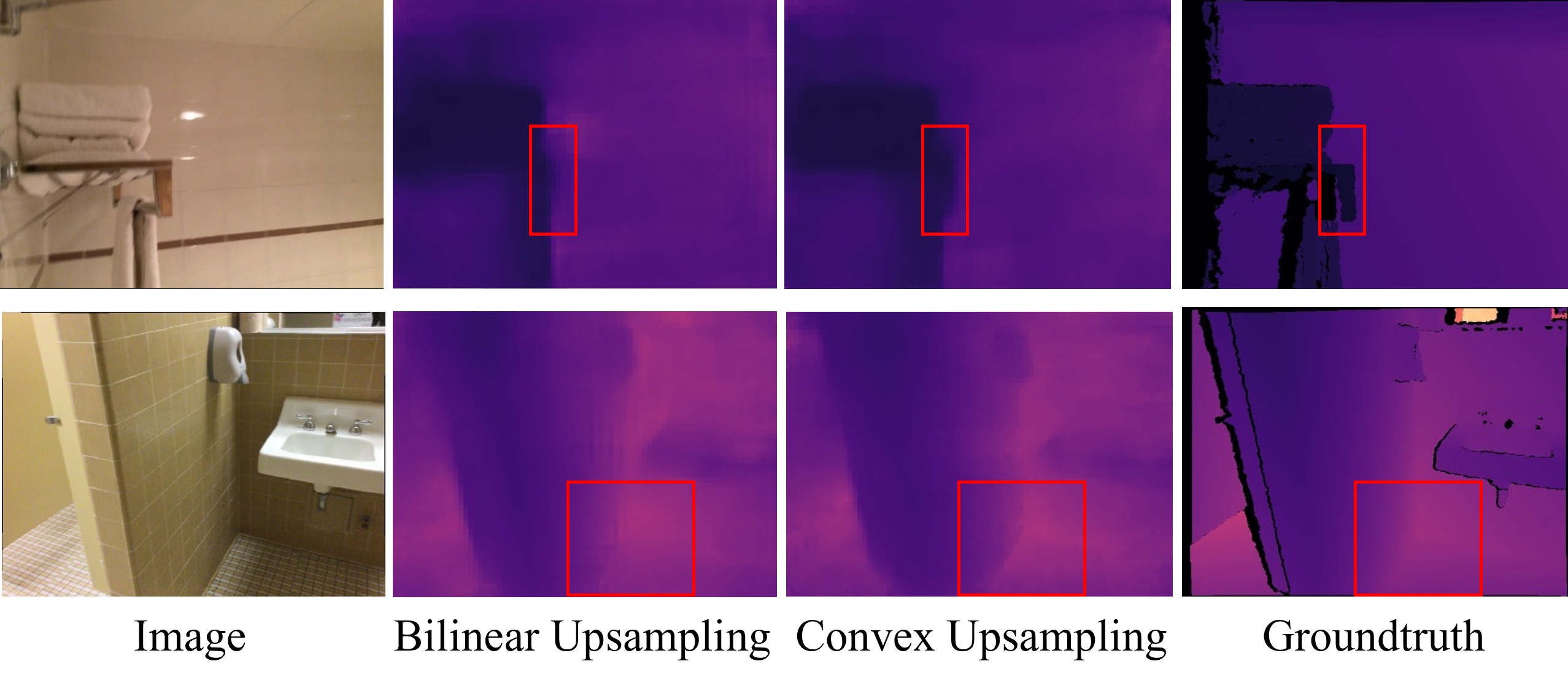}
\vspace{-0.5cm}
\caption{Effects of the convex upsampling on the depth map from pixel-level plane parameters. Regions with salient differences are highlighted with {\color{red} red} boxes. Best viewed on screen with zoom-in.}
\label{fig:convex_qualitative}
\end{figure}

This section gives some qualitative ablation analysis on the two components (\ie, convex upsampling and the soft-pooling loss) used in our method. Fig.~\ref{fig:convex_qualitative} shows the efficacy of convex upsampling. We show the depth map recovered from pixel-level parameters to eliminate the effect of plane instance pooling. It is clear that the results upsampled by convex combination have sharper boundaries and fewer artifacts than using bilinear upsampling.
\par
\begin{figure}[htp]
\flushright
\includegraphics[width=1.0\linewidth]{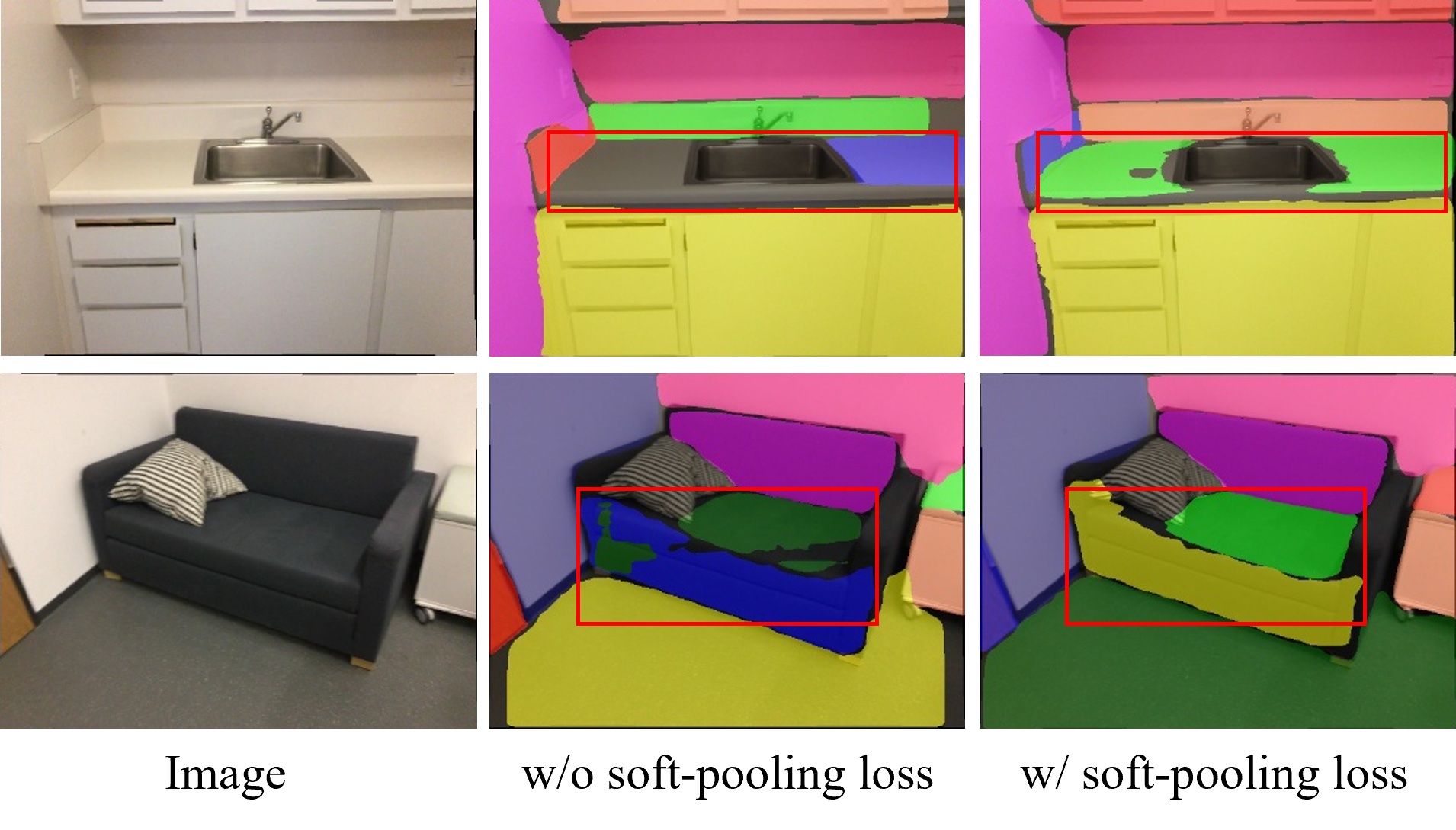}
\vspace{-0.5cm}
\caption{Effects of the soft-pooling loss on plane detection. Regions with salient differences are highlighted with {\color{red} red} boxes.}
\label{fig:soft_pooling_qualitative}
\end{figure}

Fig.~\ref{fig:soft_pooling_qualitative} shows the effectiveness of the proposed soft-pooling loss. The detected planes from the model trained with the soft-pooling loss are much more complete and align better with their boundaries.

\subsection{Additional visualizations}

\begin{figure*}[htp]
\centering
\includegraphics[width=1.0\linewidth]{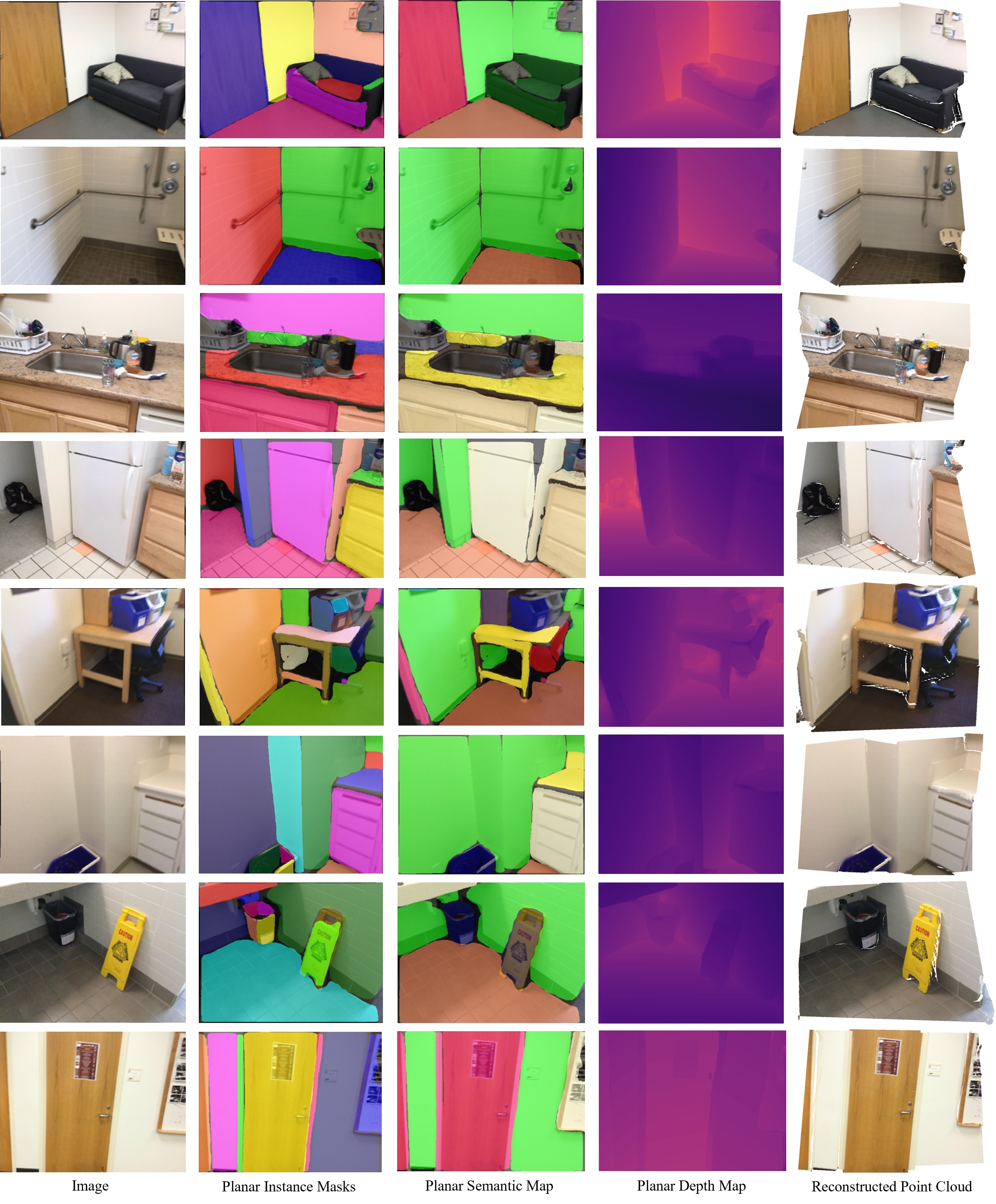}
\caption{More qualitative results on ScanNet~\cite{dai2017scannet}, including the instance planar masks, planar semantic map, planar depth map and the reconstructed 3D point cloud.}
\label{fig:scannet_more_qualitative}
\end{figure*}

We provide additional visualizations on predicted instance plane detection, planar semantic map, reconstructed planar depth map and 3D point cloud in Fig. \ref{fig:scannet_more_qualitative}, from our testing set on ScanNet~\cite{dai2017scannet}.

\subsection{Discussions and limitations}
\par
{\bf Our method \vs{}~patchmatch stereo.} Our method shares high-level ideas with traditional patchmatch stereo works~\cite{bleyer2011patchmatch,galliani2015massively} which aim to estimate a slanted plane for each pixel on the stereo reconstruction problem. However, our method differs from them in several aspects. (i)~They perform patch matching around a pixel within a squared support window, where the patch size requires to be carefully set, thus not flexible and adaptive across various real-world cases. Instead of explicitly defining a patch, we associate and match the multi-view deep features. This is based on the observation that a pixel's receptive field on the feature map is far beyond itself because of stacked CNNs. The model can automatically learn the appropriate field for matching local features with end-to-end training. (ii)~These methods usually first initialize pixels with random slanted plane hypotheses, then undergo sophisticated, multi-stage schemes with iterative optimizations. In contrast, we generate more reliable slanted plane hypotheses based on a data-driven approach (\ie, analyzing the groundtruth plane distribution), and learn the pixel-wise plane parameters in an end-to-end manner, which is much easier to optimize. (iii)~They usually adopt the photometric pixel dissimilarity as the matching cost function, which is sensitive to illumination changes and motion blurs across views. In contrast, we apply a feature-metric matching strategy, which is more robust to potential noises compared with applying photometric distance.
\par
\begin{figure}[htp]
\flushright
\includegraphics[width=1.0\linewidth]{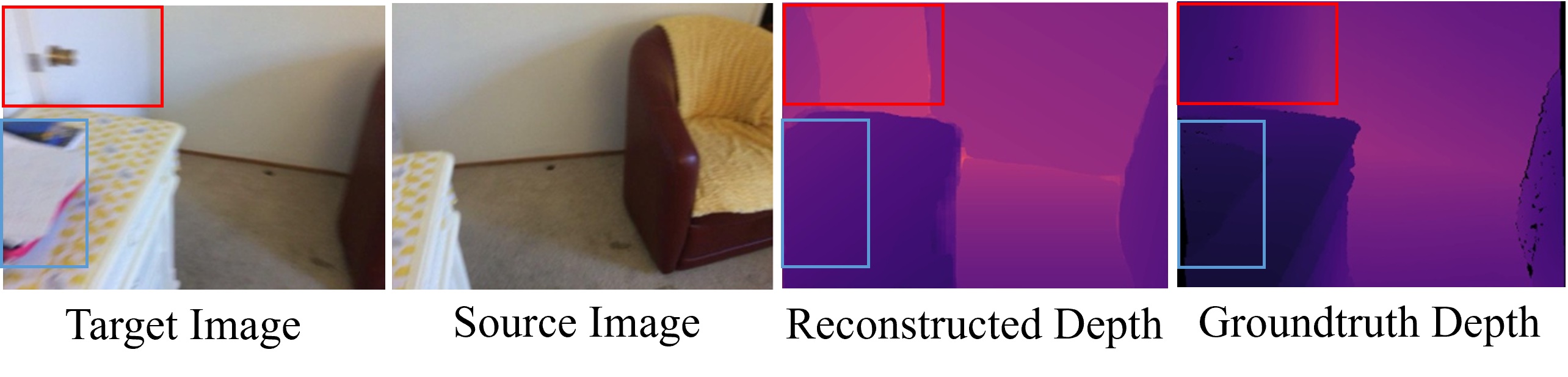}
\vspace{-0.5cm}
\caption{Failure case \uppercase\expandafter{\romannumeral1}: large temporal gap between two views. Problematic regions are highlighted with {\color{blue} blue} and {\color{red} red} boxes.}
\label{fig:failure_case_overlap}
\end{figure}
\begin{figure}[htp]
\flushright
\includegraphics[width=1.0\linewidth]{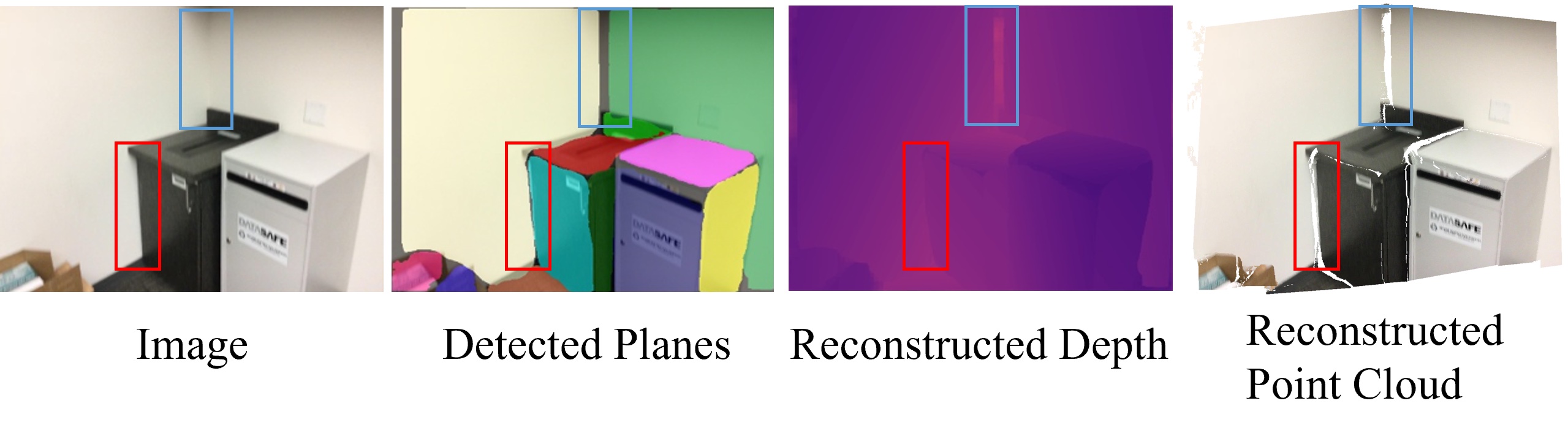}
\vspace{-0.5cm}
\caption{Failure case \uppercase\expandafter{\romannumeral2}: holes between adjacent planes. Problematic regions are highlighted with {\color{blue} blue} and {\color{red} red} boxes.}
\label{fig:failure_case_interplane}
\end{figure}

{\bf Potential limitations.}
Although we have achieved superior performance in most images, our system generates some failure cases as well. Firstly, as shown in Fig.~\ref{fig:failure_case_overlap}, because of the large temporal gap, there exist areas in the target image which are invisible in the source image and thus do not follow the planar homography relationship. This issue may be mitigated by introducing a network to learn the pixel-wise visibility or uncertainty~\cite{zhang2020visibility}. Secondly, as in Fig.~\ref{fig:failure_case_interplane}, there exist holes on some adjacent planes reconstructed from our method. An existing work~\cite{qian2020learning} proposes to infer and enforce the inter-plane relationship from single images. This approach may solve the second issue and could be incorporated to further improve the final plane reconstruction. We also leave it into future work to explore.

{\small
\bibliographystyle{ieee_fullname}
\bibliography{planemvs}
}

\end{document}